\documentclass[letterpaper]{article}
\usepackage[preprint]{aaai2027}
\nocopyright
\usepackage[hyphens]{url}
\usepackage{graphicx}
\usepackage{natbib}
\usepackage{caption}
\usepackage{amsmath,amssymb}
\usepackage{nameref}
\usepackage{multirow}
\usepackage[table]{xcolor}
\usepackage{booktabs}
\title{RiO-DETRv2: Geometry-Aware Uncertainty Modeling Meets Real-Time Oriented Object Detection Transformer}
\author{
    Zhangchi Hu\textsuperscript{1},
    Zilong He\textsuperscript{1},
    Yifan Zhao\textsuperscript{1},
    Xiangyu Yu\textsuperscript{1},
    Jie Chen\textsuperscript{1},\\
    Peixi Wu\textsuperscript{1},
    Yansong Peng\textsuperscript{1},
    Hebei Li\textsuperscript{1}\corresponding,
    Xiaoyan Sun\textsuperscript{1,2}
}
\affiliations{
    \textsuperscript{1}University of Science and Technology of China\\
    \textsuperscript{2}Institute of Artificial Intelligence, Hefei Comprehensive National Science Center
}

\date{}
\begin{document}

\maketitle

\begin{abstract}
RiO-DETR is the first real-time end-to-end transformer for oriented object detection, yet further localization gains require a more expressive box representation. Although distributional regression has succeeded in horizontal DETRs, extending it to oriented detection introduces geometry-misaligned boundary modeling and topology-inconsistent orientation refinement. We present \textbf{RiO-DETRv2}, a geometry-aware uncertainty modeling framework for real-time oriented object detection. To resolve the geometry mismatch, Object-Aligned Boundary Distribution Refinement represents box boundaries as probability distributions in an object-aligned coordinate system and progressively refines them across decoder layers. For better orientation topology consistency, we introduce Circular-to-Local
Angular Residual Refinement that transitions from global periodic search to local fine-grained estimation. Unified OBB Quality-Aligned Learning further aligns classification, boundary and angular refinement with complete-OBB localization quality. Together, these
designs improve localization without
compromising efficiency, achieving a better balance between speed and accuracy. Specifically, RiO-DETRv2-l/x achieves 53.7/54.1 mAP${_{50\text{-}95}}$ on DOTA, outperforming RiO-DETR-l/x by 1.2/1.5 points with nearly unchanged computational cost. Code is available at \url{https://github.com/RicePasteM/RiO-DETRv2}.
\end{abstract}

\section{Introduction}
\label{sec:introduction}

\begin{figure}[t] \centering \includegraphics[width=\linewidth]{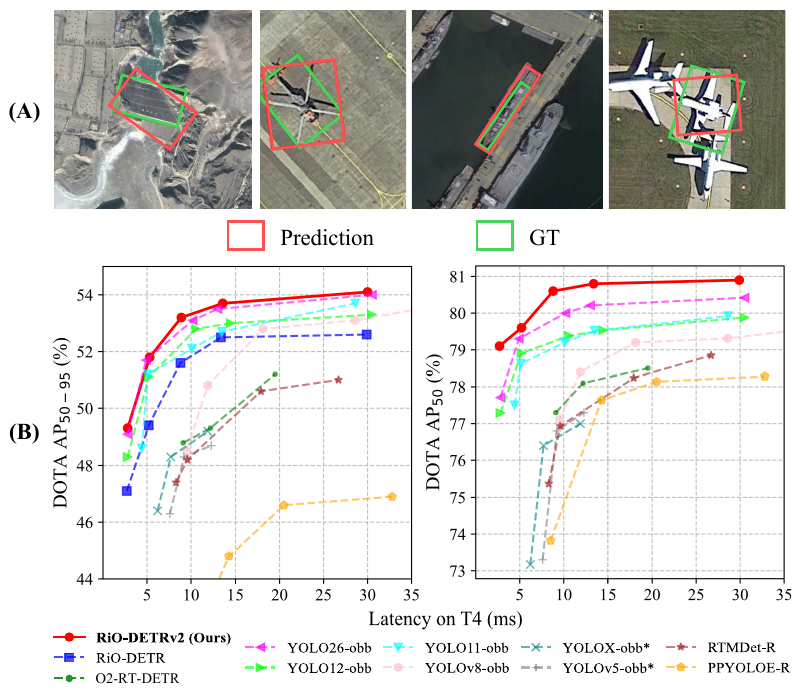} \caption{(A) Representative localization errors that are counted as TPs
under the IoU threshold of 0.5 but fail the stricter 0.75 matching criterion, revealing the need for more precise OBB localization.
(B) Comparisons with other detectors of AP$_{50\text{-}95}$ and AP$_{50}$ versus inference latency on DOTA-1.0 under single-scale training and testing protocol.}
\label{fig:teaser} 
\vspace{-1.25em}
\end{figure}

Oriented object detection extends horizontal bounding boxes to localize objects with arbitrary rotations (Oriented Bounding Boxes, OBB), which is crucial for aerial imagery, remote sensing, and scene text understanding \cite{xia2018dota, ding2021object, cheng2022anchor, zhou2022mmrotate, redmon2016you, zou2023object}. Recent real-time DETRs have demonstrated strong speed--accuracy trade-offs, and RiO-DETR \cite{hu2026rio} further extends this paradigm to oriented detection. However, the deterministic box parameterization in previous methods limits gains in localization accuracy, making it challenging to improve fine-grained OBB localization while preserving real-time efficiency.

Distributional regression has recently emerged as an effective alternative to deterministic box regression in horizontal object detection~\cite{li2020generalized,li2021generalized, peng2024d}. Instead of predicting each boundary as a single scalar, it represents possible boundary locations as a probability distribution and recovers a continuous estimate through distribution integration. Such a formulation provides richer supervision and facilitates progressive localization refinement.

However, oriented localization involves special geometric structures that are absent from horizontal bounding boxes, which makes it challenging to apply such representation. The challenges are three-fold: \textbf{(1) Geometry-misaligned boundary modeling.}
Existing distributional regressors decompose localization into four axis-aligned
boundary distributions in the global image frame. However, rotating an object
changes these regression targets even when its intrinsic geometry remains
unchanged, causing boundary uncertainty to be entangled with orientation.
\textbf{(2) Topology-inconsistent orientation refinement.} Unlike boundary distributions relative to the initial distance, angular refinement is defined in a fixed domain and 
requires both broad early-stage correction and fine-grained late-stage
estimation. A fixed support therefore suffers from a range--resolution conflict,
hindering the transition from global search to local refinement.
\textbf{(3) Quality-misaligned branch optimization.}
Although boundary and orientation should be represented independently according
to their geometric structures, they jointly determine complete-OBB localization
quality. Optimizing them with separate quality criteria may produce high
confidence predictions with unreliable overall localization.

We present \textbf{RiO-DETRv2}, a geometry-aware
uncertainty refinement framework that models spatial and angular uncertainty
according to their geometric structures while aligning optimization at the
complete-OBB level. For spatial localization, \textbf{Object-Aligned Boundary
Distribution Refinement} extends D-FINE~\cite{peng2024d} by defining four
boundary distributions in an object-aligned local coordinate system. This
representation preserves stable side semantics under arbitrary rotations,
reduces orientation-boundary entanglement, and enables progressive refinement
across decoder layers. For orientation estimation,
\textbf{Circular-to-Local Angular Residual Refinement} models layer-wise residual distributions around the current reference orientation. The first decoder layer performs topology-consistent global search over the complete periodic domain using a circular distribution, while subsequent layers employ progressively contracted local supports for fine angular estimation. The decoded residuals iteratively update the orientation through the periodic refinement mechanism of RiO-DETR, extending deterministic angle regression to coarse-to-fine uncertainty refinement.
Finally, \textbf{Unified OBB Quality-Aligned Learning} coordinates 
boundary and angular branches using complete-OBB localization quality. The quality score serves as both the soft classification target and the shared weight for distribution losses. Uncertainty statistics from predicted distributions further calibrate classification confidence, improving its consistency with the reliability of complete OBB predictions.

Extensive experiments on DOTA-1.0, DIOR-R and AI-TOD-R demonstrate that RiO-DETRv2 establishes
a better speed--accuracy trade-off for DETR-based real-time oriented object
detection. Across different model
scales and training schedules, RiO-DETRv2 consistently improves detection
accuracy over RiO-DETR with comparable computational cost. These improvements
are particularly pronounced on high-IoU-threshold metrics, which more
sensitively reflect localization precision. In particular, RiO-DETRv2-x
achieves \textbf{54.1 mAP$_{50\text{-}95}$} on DOTA-1.0 and
\textbf{53.46 mAP$_{50\text{-}95}$} / \textbf{78.32 AP$_{50}$} on DIOR-R,
while preserving real-time inference at \textbf{30.0 ms} and
\textbf{17.48 ms} on an NVIDIA T4 GPU, respectively. By unifying geometry-aware uncertainty refinement with complete-OBB quality-aligned learning, we offer a meaningful step toward more precise real-time oriented object detection, inspiring further exploration in the field.

\section{Related Works}

\subsubsection{CNN-based Oriented Object Detection.}
CNN-based oriented detectors adapt horizontal frameworks through rotated proposals, aligned RoI features, or dedicated orientation heads. Representative two-stage methods include RoI Transformer and Oriented R-CNN~\cite{ding2019learning,xie2021oriented,yang2020rotated}, while one-stage methods such as S2ANet and FCOS-based variants favor simpler and more efficient pipelines~\cite{han2021align,tian2019fcos}. Recent oriented YOLO, PP-YOLOE, and RTMDet variants further improve real-time performance~\cite{jocher2022ultralytics,ge2021yolox,xu2022pp,lyu2022rtmdet,khanam2024yolov11,tian2025yolov12,sapkota2025yolo26}, and DCFL introduces dynamic priors and coarse-to-fine assignment for challenging oriented tiny objects~\cite{xu2023dynamic}. Nevertheless, these methods generally rely on dense predictions, NMS, and rotation-specific modules, introducing additional deployment complexity and computational or memory-access overhead~\cite{cai2024poly,yuan2025strip,li2025lsknet}.

\subsubsection{DETR-based Real-Time and Oriented Detection.}
DETR formulates detection as end-to-end set prediction without handcrafted anchors or NMS~\cite{carion2020end}. Building on early DETR variants~\cite{zhu2020deformable,meng2021conditional,liu2022dab,li2022dn,zhang2022dino}, recent real-time models such as RT-DETR, LW-DETR, D-FINE, and DEIM achieve competitive efficiency through improved encoding, query initialization, distributional refinement, and supervision~\cite{zhao2024detrs,chen2024lw,peng2024d,huang2025deim,huang2025real,hu2025dome}. Oriented DETRs further introduce rotation-aware proposals, feature alignment, matching, or periodic refinement~\cite{dai2022ao2,hu2023emo2,zhao2024orientedformer,zeng2024ars,zhao2024projecting,lee2025hausdorff}. Among them, RiO-DETR establishes a real-time oriented DETR~\cite{hu2026rio}, while $\mathrm{O}^{2}$-RTDETR adapts multiple real-time DETR techniques to oriented detection~\cite{ding2026real}. However, these methods largely retain deterministic OBB parameterizations, leaving explicit spatial and angular uncertainty insufficiently explored.

\subsubsection{Angle Representation and Distributional Modeling.}
A separate line of research addresses the periodicity and boundary discontinuity of orientation prediction. CSL formulates angle estimation as circular classification with smooth labels~\cite{yang2020arbitrary}, while DCL adopts compact coded labels with angle- and aspect-ratio-aware weighting~\cite{yang2021dense}. PSC and its dual-frequency variant encode rotational periodicity through differentiable phase representations to handle boundary discontinuity and square-like ambiguity~\cite{yu2023phase}. GWD and KLD instead represent an OBB as a two-dimensional Gaussian and optimize distribution distances over center, scale, aspect ratio, and orientation~\cite{yang2021rethinking,yang2021learning}. These approaches improve periodic angle representation or construct geometry-aware regression objectives, but typically decode a deterministic OBB and do not explicitly maintain progressively refined predictive distributions for both object-aligned boundaries and periodic angular residuals.

\section{Method}
\label{sec:method}

\begin{figure*}[t]
    \centering
    \includegraphics[width=\textwidth]{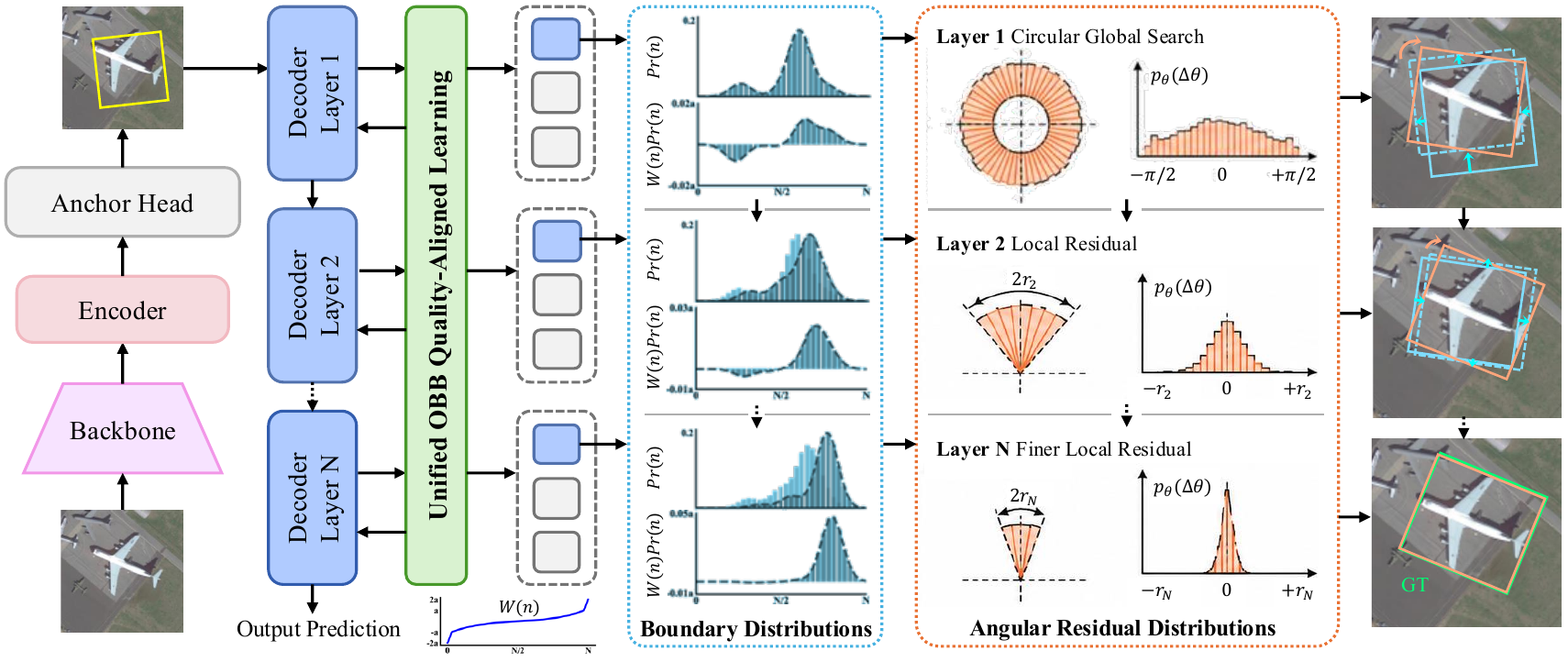}
    \caption{
        Overview of RiO-DETRv2. The OBB is factorized into object-aligned boundary geometry and periodic orientation. Side-distance distributions are progressively refined in the object frame, while the angular branch transitions from global circular search to local residual refinement. Both are jointly optimized by complete-OBB localization quality.
    }
    \label{fig:main_overview}
    \vspace{-0.5em}
\end{figure*}

We present \textbf{RiO-DETRv2}, a geometry-aware uncertainty refinement
framework for real-time oriented object detection. As illustrated in
Fig.~\ref{fig:main_overview}, RiO-DETRv2 builds upon the efficient
end-to-end architecture of RiO-DETR~\cite{hu2026rio} and introduces
geometry-aware uncertainty modeling for both spatial and angular
localization.

\subsection{Object-Aligned Boundary Distribution Refinement}
\label{sec:local_boundary_fdr}

\paragraph{Geometry misalignment of global-axis modeling.}

Distributional regression~\cite{li2020generalized, li2021generalized, peng2024d}
represents each boundary as a probability distribution over discrete candidates,
providing richer localization uncertainty modeling than deterministic regression.
For horizontal object detection, the four distributions naturally correspond to
the distances from a reference point to the top, bottom, left, and right image
boundaries, whose semantic meanings remain consistent across instances.

\begin{figure}[t] \centering \includegraphics[width=0.9\linewidth]{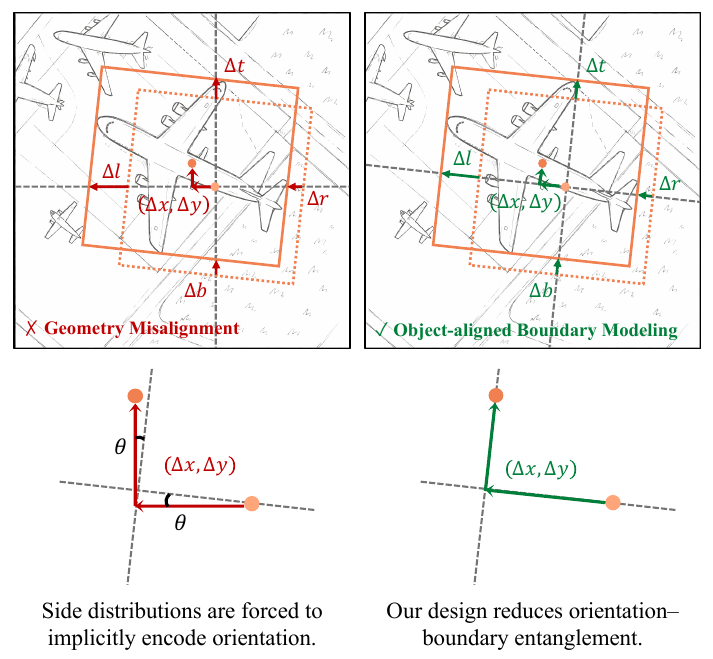} \caption{Global-axis decomposition entangles boundary distributions with orientation, whereas object-aligned modeling supervises side distances in the target-local frame and decodes them with the predicted angle, separating boundary geometry from angular estimation.} \label{fig:boundary_alignment}
\vspace{-1em}\end{figure}

However, this property does not hold for oriented objects. When an object
rotates, boundaries defined along fixed image axes no longer correspond to
consistent physical sides of the object, as shown in
Fig.~\ref{fig:boundary_alignment}. Consequently, the predicted distributions
are forced to model both intrinsic boundary uncertainty and
orientation-induced variations. Such geometry misalignment breaks side-wise
semantic consistency and limits progressive refinement across decoder layers.

\paragraph{Object-aligned boundary representation.}

To address this issue, we define boundary distributions in a canonical
coordinate frame aligned with the object geometry. Let
$\mathbf{b}^{0}=\{\mathbf{c}^{0},w^{0},h^{0},\theta^{0}\}$ and
$\mathbf{b}^{\star}=\{\mathbf{c}^{\star},w^{\star},h^{\star},\theta^{\star}\}$
denote the initial reference and target OBBs. We first transform
the center offset into the target-aligned coordinate frame:
\begin{equation}
\boldsymbol{\delta}_{\mathrm{loc}}^{\star}
=
\mathbf{R}(-\theta^{\star})
(\mathbf{c}^{\star}-\mathbf{c}^{0})
=
\begin{bmatrix}
\delta_x^{\star}\\
\delta_y^{\star}
\end{bmatrix},
\label{eq:local_center_displacement}
\end{equation}
where $\mathbf{R}(\cdot)$ denotes a two-dimensional rotation matrix. The four
physical side distances in this local frame are defined as
\begin{equation}
\mathbf{d}^{\star}
=
\begin{bmatrix}
t^{\star}\\
b^{\star}\\
\ell^{\star}\\
r^{\star}
\end{bmatrix}
=
\begin{bmatrix}
h^{\star}/2-\delta_y^{\star}\\
h^{\star}/2+\delta_y^{\star}\\
w^{\star}/2-\delta_x^{\star}\\
w^{\star}/2+\delta_x^{\star}
\end{bmatrix}.
\label{eq:object_aligned_target}
\end{equation}

Under the long-side convention, where $w\geq h$ and
$\theta\in[0,\pi)$ denotes the major-axis orientation, $\ell$ and $r$
correspond to the major-axis boundaries, while $t$ and $b$ correspond to the
minor-axis boundaries. Their physical meanings are therefore invariant to
global rotation.

During training, the target orientation $\theta^{\star}$ defines the canonical
frame, where the side-distance residuals are normalized relative to the initial
reference box and interpolated over a non-uniform candidate set. During
inference, the predicted orientation decodes the local geometry back into the
image frame. This design avoids ambiguous boundary targets caused by inaccurate
orientations and provides consistent side-wise geometric supervision.

\paragraph{Progressive boundary distribution refinement.}

The canonical object frame preserves side-wise boundary semantics across
decoder layers, enabling progressive logit accumulation. Let
$\Delta\mathbf{z}_{s}^{(l)}\in\mathbb{R}^{N}$ denote the $N$-bin logits
predicted for side $s$ at decoder layer $l$. The first layer produces the
initial logits, while subsequent layers predict additive refinements:
\begin{equation}
\mathbf{z}_{s}^{(l)}
=
\begin{cases}
\Delta\mathbf{z}_{s}^{(1)}, & l=1,\\
\mathbf{z}_{s}^{(l-1)}+\Delta\mathbf{z}_{s}^{(l)}, & l\geq2.
\end{cases}
\end{equation}
The accumulated logits are decoded into a continuous normalized residual:
\begin{equation}
\hat{\phi}_{s}^{(l)}
=
\sum_{n=0}^{N-1}
W(n)
\operatorname{Softmax}
\left(\mathbf{z}_{s}^{(l)}\right)_{n},
\label{eq:boundary_logit_refinement}
\end{equation}
for $s\in\{t,b,\ell,r\}$, where $W(n)$ is the non-uniform projection value
adopted from D-FINE~\cite{peng2024d}. The decoded residuals are converted into
layer-wise OBBs using the corresponding predicted orientations
$\theta^{(l)}$. Detailed target normalization, interpolation, and OBB decoding
are provided in Appendix~\ref{app:oabdr}.

\subsection{Circular-to-Local Angular Residual Refinement}
\label{sec:angular_refinement}

Object-aligned boundary modeling factorizes OBB localization into canonical
boundary geometry and frame orientation. The angular branch estimates the
residual from the current reference orientation to the target. Because the
initial reference may require a large periodic correction while later references
need only fine adjustment, we use a global circular support at the first layer
and progressively contracted local supports thereafter. Each layer predicts a
normalized residual distribution $\Pr_{\theta}^{(l)}(n)$, whose decoded
residual iteratively updates the orientation.

\paragraph{Circular-to-local residual refinement.}

We denote the signed shortest-path residual in the $\pi$-periodic space by
$
\left\langle x\right\rangle_{\pi}
=
\left(
(x+\pi/2)\bmod\pi
\right)
-\pi/2,
\left\langle x\right\rangle_{\pi}\in[-\pi/2,\pi/2).
$

At the first decoder layer, $M$ angular bins uniformly cover the full residual
domain, with candidates
$\psi^{(1)}(n)=-\pi/2+n\pi/M$, $n=0,\ldots,M-1$. To avoid the discontinuity
of linear decoding at the periodic boundary, we decode the first-layer residual
using the circular first moment:
\begin{equation}
\begin{aligned}
\Delta\theta^{(1)}
=
\frac{1}{2}\operatorname{atan2}
\Bigg(
&
\sum_{n=0}^{M-1}
\operatorname{Pr}\nolimits_{\theta}^{(1)}(n)
\sin\left(2\psi^{(1)}(n)\right), \\
&
\sum_{n=0}^{M-1}
\operatorname{Pr}\nolimits_{\theta}^{(1)}(n)
\cos\left(2\psi^{(1)}(n)\right)
\Bigg).
\end{aligned}
\label{eq:circular_decode}
\end{equation}
The doubled-angle representation maps the $\pi$-periodic orientation space to
a full circle, making the first and last bins adjacent. The first-layer
orientation is then updated as
$\theta^{(1)}=(\theta^{(0)}+\Delta\theta^{(1)})\bmod\pi$.

For decoder layers $l\geq2$, the residual support is progressively contracted:
\begin{equation}
\begin{aligned}
r^{(l)}
&=
\frac{\pi}{2\alpha^{l-1}},
\quad
\psi^{(l)}(n)
=
-r^{(l)}
+
\frac{2r^{(l)}n}{M-1}, \\
\Delta\theta^{(l)}
&=
\sum_{n=0}^{M-1}
\psi^{(l)}(n)
\operatorname{Pr}_{\theta}^{(l)}(n), \\
\theta^{(l)}
&=
\left(
\theta^{(l-1)}
+
\Delta\theta^{(l)}
\right)
\bmod\pi .
\end{aligned}
\label{eq:local_angle_refinement}
\end{equation}
where $\alpha>1$ is the support contraction factor. Keeping the output
dimension fixed while shrinking the support increases the effective angular
resolution of deeper layers. Consequently, the first
layer performs topology-consistent global direction search, while later layers
refine the orientation within progressively smaller local neighborhoods.

Layer $l$ is supervised by the shortest-path residual
$\phi_{\theta}^{(l)}
=\langle\theta^{\star}-\theta^{(l-1)}\rangle_{\pi}$.
The first-layer target is circularly interpolated across periodic bins, whereas
subsequent targets are clipped to the current local support and interpolated
between neighboring bins. More details are provided in Appendix~\ref{app:ctlar}.

\newcommand{\best}[1]{\textbf{#1}}
\newcommand{\second}[1]{\underline{#1}}

\begin{table*}[t]
\centering
\setlength{\tabcolsep}{2.2pt}
\renewcommand{\arraystretch}{1.00}
\setlength{\aboverulesep}{0.25ex}
\setlength{\belowrulesep}{0.25ex}
\resizebox{\textwidth}{!}{%
\begin{tabular}{@{}lccc*{15}{c}ccc@{}}
\toprule
\multirow{2}{*}{Model}
& \multirow{2}{*}{Params.}
& \multirow{2}{*}{FLOPs}
& \multirow{2}{*}{Lat.}
& \multicolumn{15}{c}{AP$_{50}$ by category}
& \multicolumn{3}{c}{Overall performance} \\
\cmidrule(lr){5-19}
\cmidrule(lr){20-22}
&
&
&
& PL & BD & BR & GTF & SV & LV & SH & TC
& BC & ST & SBF & RA & HA & SP & HC
& mAP$_{50\text{-}95}$ & AP$_{50}$ & AP$_{75}$ \\
\midrule


RTMDet-R-t
& 4.9M & 20G & 8.3
& 89.2 & 80.0 & 47.9 & 69.7 & 82.1
& 83.3 & 88.6 & 90.9 & 86.3 & 86.9
& 59.9 & 62.3 & 74.2 & 72.0 & 57.0
& 47.4 & 75.4 & 50.6 \\

YOLO26n-obb
& 2.5M & 14G & 2.8
& 89.7 & 84.7 & 52.1 & 71.1 & 83.1
& 78.9 & 88.7 & 91.1 & 87.3 & 86.9
& 58.5 & 69.1 & 76.6 & 82.8 & 65.1
& \second{49.1} & 77.4 & \best{53.4} \\

RiO-DETR-n
& 4.0M & 17G & 2.7
& 87.3 & 86.1 & 54.9
& 73.6 & 80.5
& 85.1 & 88.0 & 90.8
& 87.4 & 86.9
& 60.9 & 72.0 & 76.8
& 73.4 & 71.8
& 47.1 & \second{78.5} & 48.2 \\

\rowcolor{gray!10}
\textbf{RiO-DETRv2-n}
& 4.3M & 18G & 2.8
& 88.8 & 86.2 & 57.6
& 76.6 & 82.0
& 85.2 & 89.0 & 90.9
& 88.0 & 87.0
& 63.7 & 71.2 & 77.3
& 72.7 & 70.5
& \best{49.3} & \best{79.1} & \second{50.9} \\

\midrule


PPYOLOE-R-s
& 8.1M & 44G & 8.5
& 88.8 & 78.4 & 45.9 & 66.9 & 80.1
& 82.8 & 88.2 & 90.6 & 79.7 & 86.4
& 56.1 & 64.0 & 65.0 & 79.5 & 50.4
& 40.6 & 73.5 & 39.2 \\

RTMDet-R-s
& 8.9M & 38G & 9.7
& 89.2 & 80.5 & 52.1 & 71.4 & 81.6
& 84.1 & 88.8 & 90.9 & 87.8 & 87.0
& 59.6 & 62.3 & 75.9 & 82.0 & 61.0
& 48.2 & 76.9 & 50.6 \\

O$^{2}$-RTDETR-R18
& 20.0M & 147G & 9.1
& 89.0 & 84.2 & 52.0 & 71.5 & 80.2
& 85.0 & 88.4 & 90.9 & 87.9 & 85.5
& 57.7 & 66.3 & 77.2 & 74.6 & 69.4
& 48.8 & 77.3 & 51.1 \\

YOLO26s-obb
& 9.8M & 55G & 4.9
& 89.7 & 86.3 & 55.5
& 71.8 & 82.5
& 81.9 & 88.9 & 91.1
& 88.2 & 88.1
& 62.9 & 72.5 & 78.0
& 83.9 & 74.6
& \second{51.7} & 79.3 & \second{56.3} \\

RiO-DETR-s
& 8.2M & 53G & 5.4
& 85.6 & 84.8 & 57.5
& 75.5 & 80.7
& 86.2 & 89.0 & 90.8
& 88.4 & 88.2
& 65.2 & 74.2 & 78.3
& 80.1 & 79.6
& 49.4 & \best{79.9} & 51.0 \\

\rowcolor{gray!10}
\textbf{RiO-DETRv2-s}
& 8.9M & 53G & 5.3
& 88.5 & 86.7 & 58.5
& 77.0 & 81.6
& 85.9 & 89.1 & 90.8
& 88.0 & 88.6
& 63.1 & 71.5 & 77.9
& 76.9 & 69.5
& \best{51.8} & \second{79.6} & \best{56.4} \\

\midrule


PPYOLOE-R-m
& 24.0M & 127G & 15.3
& 89.2 & 79.8 & 51.2 & 72.9 & 81.7
& 84.4 & 88.7 & 90.8 & 86.8 & 87.5
& 58.9 & 68.0 & 73.8 & 81.7 & 68.1
& 44.8 & 77.6 & 44.7 \\

RTMDet-R-m
& 24.7M & 100G & 17.8
& 89.2 & 84.7 & 53.9 & 74.7 & 81.5
& 84.0 & 88.7 & 90.9 & 87.4 & 87.2
& 59.4 & 66.7 & 77.7 & 82.4 & 65.3
& 50.6 & 78.2 & 54.5 \\

O$^{2}$-RTDETR-R34
& 31.0M & 228G & 12.2
& 88.9 & 83.8 & 53.8 & 73.9 & 80.8
& 85.4 & 88.6 & 90.9 & 86.6 & 86.8
& 58.3 & 66.4 & 77.3 & 80.1 & 70.7
& 49.3 & 78.1 & 50.7 \\

YOLO26m-obb
& 21.2M & 183G & 10.2
& 89.2 & 86.8 & 56.4
& 73.0 & 82.6
& 78.3 & 89.1 & 91.2
& 88.4 & 87.9
& 64.5 & 78.0 & 78.3
& 83.1 & 79.7
& \second{53.1} & 79.3 & \best{58.4} \\

RiO-DETR-m
& 18.6M & 158G & 8.8
& 85.9 & 86.4 & 61.7
& 78.1 & 82.0
& 86.8 & 89.2 & 90.8
& 88.9 & 87.7
& 67.4 & 73.3 & 78.8
& 77.4 & 78.8
& 51.6 & \second{80.6} & 55.0 \\

\rowcolor{gray!10}
\textbf{RiO-DETRv2-m}
& 19.5M & 158G & 8.9
& 85.6 & 87.5 & 60.0
& 80.3 & 81.7
& 86.3 & 89.3 & 90.9
& 88.3 & 88.7
& 69.3 & 72.4 & 78.8
& 81.6 & 77.1
& \best{53.2} & \best{80.8} & \second{57.3} \\

\midrule


PPYOLOE-R-l
& 53.3M & 282G & 20.5
& 89.2 & 81.0 & 54.0 & 70.2 & 81.9
& 85.2 & 88.8 & 90.8 & 87.0 & 88.0
& 62.9 & 67.9 & 76.6 & 79.1 & 69.7
& 46.6 & 78.1 & 48.5 \\

RTMDet-R-l
& 52.3M & 205G & 26.7
& 89.4 & 84.2 & 55.2 & 75.1 & 80.8
& 84.5 & 89.0 & 90.9 & 87.4 & 87.3
& 63.1 & 67.9 & 78.1 & 80.8 & 69.1
& 51.0 & 78.9 & 55.2 \\

O$^{2}$-RTDETR-R50
& 42.0M & 339G & 19.5
& 89.3 & 84.1 & 56.5 & 73.9 & 80.0
& 84.1 & 88.1 & 90.9 & 85.3 & 86.1
& 63.2 & 65.0 & 77.8 & 77.8 & 72.4
& 51.2 & 78.5 & 54.3 \\

YOLO26l-obb
& 25.6M & 230G & 13.0
& 89.5 & 86.3 & 56.2
& 74.2 & 82.9
& 78.8 & 89.3 & 91.1
& 88.6 & 88.1
& 67.9 & 72.2 & 78.2
& 83.5 & 76.5
& \second{53.5} & 79.8 & \best{59.1} \\

RiO-DETR-l
& 27.5M & 230G & 13.4
& 86.7 & 86.8 & 60.8
& 79.4 & 82.0
& 86.4 & 89.0 & 90.8
& 88.7 & 88.5
& 70.0 & 76.0 & 79.2
& 83.4 & 76.9
& 52.5 & \best{81.5} & 56.2 \\

\rowcolor{gray!10}
\textbf{RiO-DETRv2-l}
& 28.6M & 230G & 13.6
& 86.9 & 86.8 & 60.7
& 76.7 & 82.7
& 86.1 & 89.3 & 90.9
& 88.2 & 88.4
& 71.4 & 71.7 & 79.0
& 83.0 & 69.9
& \best{53.7} & \second{80.8} & \second{58.7} \\

\midrule


PPYOLOE-R-x
& 100M & 530G & 32.8
& 89.5 & 79.7 & 55.0 & 75.6 & 82.4
& 85.2 & 88.4 & 90.8 & 85.7 & 87.7
& 63.2 & 69.5 & 77.1 & 75.1 & 69.4
& 46.9 & 78.3 & 49.3 \\

YOLO26x-obb
& 57.6M & 517G & 30.5
& 89.7 & 86.6 & 57.3
& 73.8 & 82.8
& 79.3 & 89.3 & 91.1
& 89.1 & 88.1
& 66.7 & 70.6 & 77.9
& 83.3 & 80.7
& \second{54.0} & 79.8 & \second{59.3} \\

RiO-DETR-x
& 62.5M & 527G & 29.9
& 88.3 & 87.5 & 61.9
& 79.2 & 83.2
& 86.5 & 89.2 & 90.9
& 88.4 & 87.8
& 71.2 & 75.1 & 78.5
& 82.2 & 77.1
& 52.6 & \best{81.8} & 57.6 \\

\rowcolor{gray!10}
\textbf{RiO-DETRv2-x}
& 64.6M & 528G & 30.0
& 87.7 & 86.3 & 62.5
& 79.0 & 81.0
& 85.3 & 88.9 & 90.8
& 87.0 & 88.4
& 71.5 & 70.6 & 78.1
& 80.3 & 77.3
& \best{54.1} & \second{80.9} & \best{59.4} \\

\bottomrule
\end{tabular}%
}
\caption{
Comparison with representative real-time oriented object detectors on the
DOTA-1.0 test set. For each matched model scale, the best and second-best
overall results are highlighted in \textbf{bold} and
\underline{underlined}, respectively.
}
\label{tab:dota_main_results}
\vspace{-0.5em}
\end{table*}

\begin{table}[t]
\centering

\scriptsize
\setlength{\tabcolsep}{2.3pt}
\renewcommand{\arraystretch}{0.92}

\setlength{\aboverulesep}{0.25ex}
\setlength{\belowrulesep}{0.25ex}

\resizebox{\columnwidth}{!}{%
\begin{tabular}{@{}lcccccc@{}}
\toprule
Model
& Para.
& FLOPs
& Lat.
& mAP$_{50\text{-}95}$
& AP$_{50}$
& AP$_{75}$ \\
\midrule

YOLO26n-obb
& 2.5M & 10G & 2.23
& \second{45.79}
& 71.88
& \second{46.92} \\

RiO-DETR-n
& 4.0M & 11G & 2.20
& 44.06
& \second{71.92}
& 44.57 \\

\rowcolor{gray!10}
\textbf{RiO-DETRv2-n}
& 4.3M & 11G & 2.27
& \best{46.47}
& \best{72.21}
& \best{47.69} \\

\midrule

YOLO26s-obb
& 9.8M & 39G & 2.84
& \second{48.41}
& 74.07
& \best{51.30} \\

RiO-DETR-s
& 8.2M & 33G & 3.01
& 47.94
& \second{74.44}
& 49.09 \\

\rowcolor{gray!10}
\textbf{RiO-DETRv2-s}
& 8.9M & 33G & 3.29
& \best{49.95}
& \best{75.13}
& \second{51.29} \\

\midrule

O$^{2}$-RTDETR-R34
& 31M & 141G & 8.20
& 45.12
& 68.70
& 45.70 \\

YOLO26m-obb
& 21.2M & 129G & 5.91
& \second{50.56}
& 74.65
& \second{53.35} \\

RiO-DETR-m
& 18.6M & 97G & 5.10
& 48.03
& \second{75.73}
& 49.46 \\

\rowcolor{gray!10}
\textbf{RiO-DETRv2-m}
& 19.5M & 98G & 5.35
& \best{52.61}
& \best{77.28}
& \best{54.28} \\

\midrule

O$^{2}$-RTDETR-R50
& 42M & 209G & 12.41
& 47.79
& 72.20
& 48.20 \\

YOLO26l-obb
& 25.6M & 158G & 7.53
& \second{51.15}
& 75.31
& 51.87 \\

RiO-DETR-l
& 27.5M & 141G & 7.76
& 50.64
& \second{76.11}
& \second{52.10} \\

\rowcolor{gray!10}
\textbf{RiO-DETRv2-l}
& 28.6M & 142G & 8.13
& \best{52.85}
& \best{77.57}
& \best{54.45} \\

\midrule

YOLO26x-obb
& 57.6M & 353G & 17.66
& \second{52.28}
& 76.48
& \second{55.11} \\

RiO-DETR-x
& 62.5M & 324G & 17.31
& 51.80
& \second{77.43}
& 53.82 \\

\rowcolor{gray!10}
\textbf{RiO-DETRv2-x}
& 64.6M & 325G & 17.48
& \best{53.46}
& \best{78.32}
& \best{55.33} \\

\bottomrule
\end{tabular}
}
\caption{Performance comparison with state-of-the-art real-time oriented object detectors on DIOR-R.}
\label{tab:dior_r_results}
\end{table}

\begin{table}[t]
\centering
\scriptsize
\setlength{\tabcolsep}{2.1pt}
\renewcommand{\arraystretch}{0.95}
\setlength{\aboverulesep}{0.25ex}
\setlength{\belowrulesep}{0.25ex}
\resizebox{\columnwidth}{!}{%
\begin{tabular}{@{}lcccccc@{}}
\toprule
Model
& Params.
& FLOPs
& Lat.
& mAP$_{50\text{-}95}$
& AP$_{50}$
& AP$_{75}$ \\
\midrule

Oriented R-CNN$^{\dagger}$
& 41.1M & 297G & --
& 11.2 & 33.2 & 4.3 \\

$\mathrm{S}^{2}\mathrm{A}$-Net$^{\dagger}$
& 38.6M & 271G & --
& 10.8 & 33.4 & 3.3 \\

Oriented RepPoints$^{\dagger}$
& 36.6M & 269G & --
& 13.0 & 40.3 & 4.2 \\

ARS-DETR$^{\dagger}$
& 41.1M & 250G & 25.32
& 14.3 & 41.1 & 5.8 \\

DCFL ($\mathrm{S}^{2}\mathrm{A}$-Net)$^{\dagger}$
& 38.6M & 271G & --
& 17.5 & 49.6 & 7.9 \\

RHINO-DETR
& 50.8M & 383G & 35.56
& 17.7 & 53.9 & 5.7 \\

\midrule

YOLO26m-obb
& 21.2M & 97G & 5.10
& \second{19.8}
& \best{56.5}
& \second{8.5} \\

RiO-DETR-m
& 18.6M & 102G & 6.51
& 18.2
& 54.5
& 6.2 \\

\rowcolor{gray!10}
\textbf{RiO-DETRv2-m}
& 19.5M & 103G & 6.70
& \best{21.0}
& \second{55.7}
& \best{10.4} \\

\midrule

YOLO26x-obb
& 57.6M & 324G & 17.31
& \second{19.9}
& \second{56.8}
& \best{8.8} \\

RiO-DETR-x
& 62.5M & 336G & 21.03
& 18.6
& 55.2
& 6.6 \\

\rowcolor{gray!10}
\textbf{RiO-DETRv2-x}
& 64.6M & 339G & 21.23
& \best{20.6}
& \best{59.2}
& \second{8.6} \\

\bottomrule
\end{tabular}%
}
\caption{
Comparison with state-of-the-art oriented object detectors on the AI-TOD-R benchmark. 
$\dagger$ denotes results reported by the official benchmark
\cite{xu2024orientedtinyobjectdetection}.
}
\label{tab:aitodr_results}
\end{table}


\subsection{Unified OBB Quality-Aligned Learning}
\label{sec:quality_learning}

We introduce Unified OBB Quality-Aligned Learning to coordinate classification,
boundary refinement, and angular refinement using the localization quality of
the complete oriented box. It further calibrates classification confidence
using uncertainty statistics extracted from the predicted boundary and angular
distributions.

\paragraph{Unified OBB quality supervision.}

For the $k$-th matched prediction at decoder layer $l$, we convert the Gaussian KLD loss~\cite{yang2021learning} into a bounded localization quality:
\begin{equation}
q_k^{(l)}
=
\max
\left(
1-
\mathcal{L}_{\mathrm{KLD}}
\left(
\mathbf{b}_k^{(l)},
\mathbf{b}_k^{\star}
\right),
0
\right).
\label{eq:obb_quality}
\end{equation}
We detach $q_k^{(l)}$ from the computation graph. As a complete-OBB measure,
it reflects the joint effects of center, scale, aspect-ratio, and orientation
errors. We use it as the positive soft target in Varifocal Loss. The same quality also
weights both distribution refinement objectives:
\begin{equation}
\mathcal{L}_{\mathrm{FGL}}
=
\frac{1}{K_m}
\sum_{l=1}^{L}
\sum_{k=1}^{K_m}
q_k^{(l)}
\left(
\ell_{\mathrm{B},k}^{(l)}
+
\ell_{\mathrm{A},k}^{(l)}
\right),
\label{eq:quality_weighted_fgl}
\end{equation}
where $\ell_{\mathrm{B}}$ and $\ell_{\mathrm{A}}$ denote the boundary and
angular distribution refinement losses, respectively; $K_m$ is the number of
matched queries, and $L$ is the number of decoder layers.

\paragraph{Distribution-aware confidence calibration.}

Unified OBB quality aligns classification confidence with complete-box
localization quality, while the predicted distributions provide complementary
uncertainty information. For the boundary branch, we extract confidence
statistics from the predicted boundary distributions and use them to calibrate
the classification logit through a lightweight MLP.

For the angular branch, we use normalized entropy to measure predictive
uncertainty independent of the layer-wise support radius:
\begin{equation}
\mathcal{H}_{\theta}^{(l)}
=
-
\frac{
\sum_{n=0}^{M-1}
p_{\theta}^{(l)}(n)
\log p_{\theta}^{(l)}(n)
}{
\log M
}.
\label{eq:angular_entropy}
\end{equation}

Orientation reliability also depends on the intrinsic shape of the predicted
box. Near-square objects have ambiguous principal directions even when their
angular distributions are concentrated. We therefore define the anisotropy of
the final decoded box as
\begin{equation}
a^{(L)}
=
1-
\frac{
\min(w^{(L)},h^{(L)})
}{
\max(w^{(L)},h^{(L)})
}.
\label{eq:box_anisotropy}
\end{equation}

Angular entropy captures predictive uncertainty, whereas anisotropy captures intrinsic identifiability of orientation. We concatenate layer-wise
angular uncertainty statistics and final-box anisotropy with boundary
distribution statistics, and use a lightweight MLP to calibrate the
classification logit. Detailed formulations are provided in
Appendix~\ref{app:calibration}.

\section{Experiments}

In this section, we evaluate RiO-DETRv2 through comparisons with state-of-the-art
methods and comprehensive ablation studies. We further analyze the underlying
mechanisms of the proposed method in Appendices~\ref{sec:appendix_uncertainty}--\ref{app:near_square}.

\subsection{Evaluation Datasets}

\subsubsection{DOTA-1.0.}

DOTA-1.0 \cite{xia2018dota} is a large-scale oriented object detection dataset containing 2,806 images and 188,282 instances across 15 categories, with significant variations in orientation, shape, and scale. We use single-scale setting for evaluation, images are cropped into $1024 \times 1024$ patches with a stride of 824. We
submit test predictions to the official evaluation server to obtain final performance.

\subsubsection{DIOR-R.}

DIOR-R \cite{cheng2022anchor} is constructed from DIOR by adding annotations for rotated bounding boxes. It comprises 23,463 images of size $800 \times 800$, totaling 190,288 oriented instances across 20 categories. The dataset is split into training, validation, and testing sets with a ratio of 1:1:2.

\subsubsection{AI-TOD-R.}

AI-TOD-R \cite{xu2024orientedtinyobjectdetection} is a large-scale benchmark for oriented tiny object detection, containing 28,036 images and 752,460 oriented instances across eight categories. Each image has a resolution of $800 \times 800$, while the mean object size is only approximately $10.6$ pixels. The dataset presents substantial challenges arising from extremely small object scales, arbitrary orientations, dense object arrangements, and imbalanced category distributions.

\begin{figure*}[t]
    \centering
    \includegraphics[width=\textwidth]{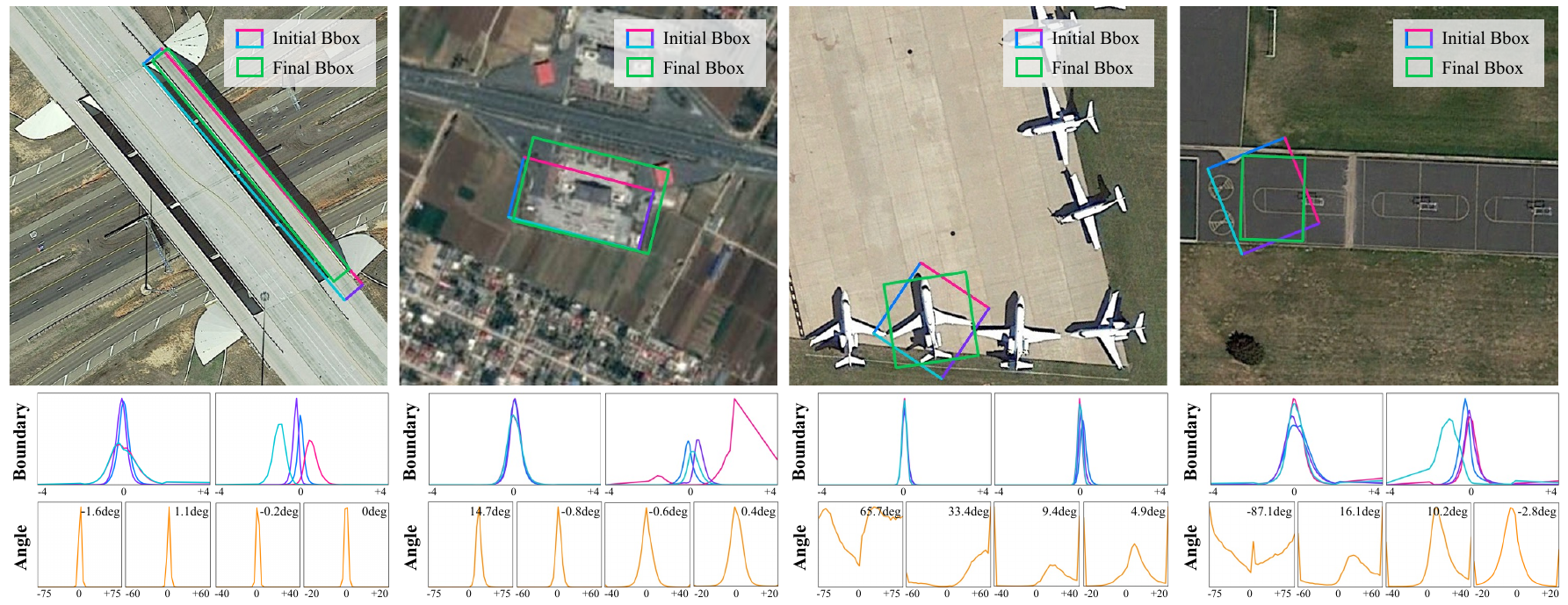}
    \caption{
Visualization of geometry-aware uncertainty refinement. The top row compares the initial and final OBBs, while the lower rows show layer-wise boundary and angular distributions. Boundary predictions progressively concentrate in the object-aligned frame, and angular refinement transitions from large periodic corrections to fine local adjustments.
    }
    \label{fig:vis_dis}
\end{figure*}

\subsection{Implementation Details}
For all datasets, models are trained on the \textit{trainval} set and evaluated on the \textit{test} set following their official protocols. Results of DOTA are returned by the official server. All models are trained on 2 NVIDIA A800 or 4 RTX3090 GPUs. All latencies are measured equally on a single NVIDIA T4 GPU using TensorRT 10 with FP16 following the same setting of RiO-DETR~\cite{hu2026rio}. More implementation details are provided in Appendix~\ref{app:implementation}. 

\subsection{Comparison with State-of-the-Art Methods}
To evaluate our proposed RiO-DETRv2, we compare with these state-of-the-art methods: RTMDet-R~\cite{lyu2022rtmdet}, PPYOLOE-R~\cite{xu2022pp}, YOLO26-obb~\cite{sapkota2025yolo26}, RiO-DETR~\cite{hu2026rio}, $\text{O}^2$-RTDETR~\cite{ding2026real}, Oriented R-CNN~\cite{xie2021oriented}, $\text{S}^2\text{A}$-Net~\cite{han2021align}, Oriented RepPoints~\cite{li2022oriented}, ARS-DETR~\cite{zeng2024ars}, DCFL~\cite{xu2023dynamic,xu2024orientedtinyobjectdetection} and RHINO-DETR~\cite{lee2025hausdorff}.

\subsubsection{Results on DOTA-1.0.}
As shown in Table~\ref{tab:dota_main_results}, RiO-DETRv2 consistently improves localization accuracy over RiO-DETR across all model scales with nearly unchanged computational cost. Specifically, RiO-DETRv2-n/s/m/l/x achieves 49.3/51.8/53.2/53.7/54.1 mAP$_{50\text{-}95}$, outperforming the corresponding baselines by 2.2/2.4/1.6/1.2/1.5 points with only marginal increases in parameters and FLOPs. The consistent AP$_{75}$ gains further confirm improved fine-grained OBB localization. Under comparable computational budgets, RiO-DETRv2 also surpasses real-time competitors in mAP$_{50\text{-}95}$ at all scales while retaining competitive AP$_{50}$. Slight AP$_{50}$ variations at some scales may arise from quality-aware confidence calibration, which downweights predictions with insufficient complete-OBB localization quality and changes ranking behavior under loose IoU thresholds.

\subsubsection{Results on DIOR-R.}
Table~\ref{tab:dior_r_results} demonstrates the generalization ability
of RiO-DETRv2 on DIOR-R. Compared with both YOLO26-obb and the original
RiO-DETR, RiO-DETRv2 consistently achieves the highest
mAP$_{50\text{-}95}$ across all model scales. In particular,
RiO-DETRv2-x reaches 53.46 mAP$_{50\text{-}95}$ and 78.32 AP$_{50}$,
outperforming YOLO26x-obb by 1.18 and 1.84 points while
maintaining similar computational cost and latency. The improvements become
more evident at AP$_{75}$, where RiO-DETRv2 provides more accurate oriented
box localization under stricter IoU requirements.

\subsubsection{Results on AI-TOD-R.}
AI-TOD-R is challenging due to its tiny, densely distributed, and arbitrarily oriented objects, for which small boundary or angular errors can cause substantial IoU degradation. As shown in Table~\ref{tab:aitodr_results}, RiO-DETRv2-m improves mAP$_{50\text{-}95}$ from 18.2 to 21.0 and AP$_{75}$ from 6.2 to 10.4 over RiO-DETR, while RiO-DETRv2-x gains 2.0 mAP$_{50\text{-}95}$ and 4.0 AP$_{50}$. Under comparable computational budgets, RiO-DETRv2 also outperforms YOLO26-obb in mAP$_{50\text{-}95}$ at both scales and achieves a notable AP$_{75}$ gain at the m scale, demonstrating the benefit of geometry-aware uncertainty refinement for OBB localization.

\begin{table}[t]
\centering

\setlength{\tabcolsep}{3.2pt}
\renewcommand{\arraystretch}{1.05}

\resizebox{\columnwidth}{!}{%
\begin{tabular}{@{}lccc cccc@{}}
\toprule
Dataset
& OABDR
& CTLAR
& UniOBB-QAL
& Params.
& FLOPs
& AP$_{50\text{-}95}$
& AP$_{50}$ \\
\midrule

\multirow{5}{*}{DOTA-1.0}
&
&
&
& 18.6M
& 158G
& 51.6
& 80.6
\\

&
\checkmark
&
&
& 19.5M
& 158G
& 52.2
& 81.0
\\

&
&
\checkmark
&
& 18.7M
& 158G
& 51.9
& \second{81.1}
\\

&
\checkmark
& \checkmark
&
& 19.5M
& 158G
& \second{52.7}
& \best{81.2}
\\

\rowcolor{gray!10}
&
\checkmark
& \checkmark
& \checkmark
& 19.5M
& 158G
& \best{53.2}
& 80.8
\\

\midrule

\multirow{5}{*}{DIOR-R}
&
&
&
& 18.6M
& 97G
& 48.0
& 75.7
\\

&
\checkmark
&
&
& 19.5M
& 98G
& 49.6
& 76.5
\\

&
&
\checkmark
&
& 18.7M
& 97G
& 49.4
& 76.8
\\

&
\checkmark
& \checkmark
&
& 19.5M
& 98G
& \second{51.2}
& \best{77.4}
\\

\rowcolor{gray!10}
&
\checkmark
& \checkmark
& \checkmark
& 19.5M
& 98G
& \best{52.6}
& \second{77.3}
\\

\bottomrule
\end{tabular}%
}
\caption{
Component ablation study of RiO-DETRv2-m.
}
\label{tab:ablation_main}
\vspace{-1em}
\end{table}

\begin{table}[t]
\centering

\setlength{\tabcolsep}{4pt}
\renewcommand{\arraystretch}{1.05}

\resizebox{\columnwidth}{!}{%
\begin{tabular}{@{}lccc@{}}
\toprule
Boundary Representation (on DIOR-R)
& AP$_{50\text{-}95}$
& AP$_{50}$
& AP$_{75}$ \\
\midrule

Original RiO-DETR (Base)
& 51.80
& \second{77.43}
& 53.82
\\

w/ Global-axis distribution
& \second{52.19}
& 77.22
& \second{54.20}
\\

\rowcolor{gray!10}
\textbf{w/ Object-aligned distribution (Ours)}
& \textbf{52.88}
& \textbf{77.74}
& \textbf{54.66}
\\

\bottomrule
\end{tabular}
}
\caption{
Ablation study on OABDR of RiO-DETRv2-x.
}
\label{tab:ablation_oabdr}
\end{table}

\begin{table}[t]
\centering

\setlength{\tabcolsep}{4pt}
\renewcommand{\arraystretch}{1.05}

\resizebox{\columnwidth}{!}{%
\begin{tabular}{@{}lccc@{}}
\toprule
Angular Refinement (on DIOR-R)
& AP$_{50\text{-}95}$
& AP$_{50}$
& AP$_{75}$ \\
\midrule

Original RiO-DETR w/ OABDR (Base)
& 52.88
& 77.74
& 54.66
\\

w/ Linear angular distribution
& 52.66
& 77.31
& 53.82
\\

w/ Circular distribution
& 53.02
& 78.14
& 54.78
\\

\rowcolor{gray!10}
\textbf{w/ Circular-to-local refinement (Ours)}
& \textbf{53.16}
& \textbf{78.39}
& \textbf{55.13}
\\

\bottomrule
\end{tabular}
}
\caption{
Ablation study on CTLAR of RiO-DETRv2-x.
}
\label{tab:ablation_ctlar}
\vspace{-1em}
\end{table}

\subsection{Ablation Studies}

\subsubsection{Ablation on Main Components.}

As shown in Table~\ref{tab:ablation_main}, both Object-Aligned Boundary
Distribution Refinement (OABDR) and Circular-to-Local Angular Residual
Refinement (CTLAR) consistently improve performance on DOTA-1.0 and DIOR-R.
Combining them further increases AP$_{50\text{-}95}$ from 51.6 to 52.7 on
DOTA-1.0 and from 48.0 to 51.2 on DIOR-R, validating their complementary
effects on boundary and angular refinement. Unified OBB Quality-Aligned
Learning (UniOBB-QAL) further improves AP$_{50\text{-}95}$ to 53.2 and 52.6,
respectively, with negligible overhead. The slight AP$_{50}$ decrease on DOTA-1.0 results from improved confidence calibration, where predictions with lower complete-OBB quality receive reduced confidence, leading to more reliable high-IoU localization.

\subsubsection{Ablation on OABDR.}
As shown in Table~\ref{tab:ablation_oabdr}, replacing the object-aligned
boundary representation with global-axis distributions only improves the
baseline slightly, achieving 52.19 AP$_{50\text{-}95}$ compared with 51.80
of the original RiO-DETR. In contrast, introducing the proposed
object-aligned distribution increases AP$_{50\text{-}95}$ to 52.88 and
AP$_{75}$ to 54.66. This demonstrates that the improvement of OABDR mainly
comes from establishing geometry-consistent boundary representations, which
preserve stable side semantics under arbitrary object rotations rather than
simply applying distributional regression to OBBs.

\subsubsection{Ablation on CTLAR.}
As shown in Table~\ref{tab:ablation_ctlar}, using a linear angular distribution slightly degrades the performance, reducing AP$_{50\text{-}95}$ from 52.88 to 52.66. In contrast, the circular distribution improves angular modeling by explicitly accounting for the periodicity of orientations, increasing AP$_{50\text{-}95}$ to 53.02 and AP$_{50}$ to 78.14. Building upon circular representation, the proposed circular-to-local angular refinement further improves the performance to 53.16 AP$_{50\text{-}95}$, 78.39 AP$_{50}$, and 55.13 AP$_{75}$. Compared with circular decoding alone, it yields gains of 0.14, 0.25, and 0.35 points, respectively. These results demonstrate combining topology-consistent global angular modeling with progressive local refinement enables more accurate angular estimation.

\subsection{Visualization Analysis}
Fig.~\ref{fig:vis_dis} shows that RiO-DETRv2 progressively refines
coarse initial OBBs into more accurate final predictions. The boundary
distributions become concentrated around the refined object-aligned side
distances, confirming the effectiveness of geometry-consistent boundary
modeling. Meanwhile, the angular residuals exhibit a coarse-to-fine pattern:
early layers perform large periodic corrections, while later layers predict
small local residuals within contracted supports. This demonstrates that the
proposed refinement framework improves both boundary localization and angular
precision.

\section{Conclusion}
We present \textbf{RiO-DETRv2}, a geometry-aware uncertainty refinement framework for real-time end-to-end oriented object detection. It models boundaries in an object-aligned frame, refines orientation through circular-to-local residual distributions, and aligns classification and localization using complete-OBB quality. Experiments on DOTA-1.0, DIOR-R, and AI-TOD-R demonstrate consistent localization gains with comparable computational cost, establishing a better speed--accuracy trade-off for real-time oriented object detection. \textbf{Limitation and Future Work:} We further observe an increasingly pronounced trade-off among evaluation metrics between recall and precision as detector performance improves in the field of oriented object detection.
Future work may focus on balancing overall detection quality and developing
more informative evaluation protocols. We also encourage future studies to
report comprehensive threshold-based metrics alongside conventional metrics to
provide a more holistic assessment of detector performance.

\bibliography{references}

\clearpage
\appendix
\setcounter{secnumdepth}{2}

\section{Additional Details of OABDR}
\label{app:oabdr}

This section provides target assignment, inverse normalization, and layer-wise
OBB decoding details of Object-Aligned Boundary Distribution Refinement.

\subsection{Boundary Target Assignment}

For each object-aligned side $s\in\mathcal{S}=\{t,b,\ell,r\}$, the normalized
side residual target $\phi_s^\star$ is projected onto the non-uniform candidate
set $\{W(n)\}_{n=0}^{N-1}$. Two adjacent bins are selected and interpolation
weights are computed for distribution supervision.

\subsection{Inverse Normalization}

The continuous residual predicted at decoder layer $(l)$ is

\[
\hat{\boldsymbol{\phi}}^{(l)}
=
[\hat{\phi}_{t}^{(l)},\hat{\phi}_{b}^{(l)},
\hat{\phi}_{\ell}^{(l)},\hat{\phi}_{r}^{(l)}]^{\mathsf T}.
\]

The refined physical side distances are recovered as

\[
\mathbf d^{(l)}
=
\mathbf d^0+
\frac{\mathbf s^0}{\rho}
\odot
\hat{\boldsymbol{\phi}}^{(l)},
\]

where $\mathbf d^0$, $\mathbf s^0$, and $\rho$ denote the reference side
distances, normalization factors, and regression scale used for target
construction.

\subsection{Layer-wise OBB Decoding}

The refined side distances are decoded in the canonical frame and transformed
back using the predicted orientation $\theta^{(l)}$.

\section{Additional Details of CTLAR}
\label{app:ctlar}

For a matched target orientation $\theta^\star$, the residual target at decoder
layer $(l)$ is

\[
\phi_\theta^{(l)}
=
\langle\theta^\star-\theta^{(l-1)}\rangle_\pi .
\]

The first layer uses circular interpolation over the full periodic domain,
while later layers use clipped local supports.

The angular distribution loss is

\[
\ell_{\mathrm A,k}^{(l)}
=
\sum_j\omega_{j,k}^{(l)}
\operatorname{CE}
(p_{\theta,k}^{(l)},n_{j,k}^{(l)}).
\]

\section{Distribution-aware Confidence Calibration}
\label{app:calibration}

Let $\mathbf g^{(l)}$ denote the classification logit at decoder layer $(l)$.
For the boundary branch, concentration statistics are extracted from the
predicted object-aligned boundary distributions. The implementation uses the
top-$K$ probabilities of each side distribution as a compact descriptor:

\[
\mathbf v_s^{(l)}
=
\operatorname{TopK}(p_s^{(l)},K).
\]

The boundary descriptor is

\[
\mathbf u_B^{(l)}
=
\operatorname{Concat}_{s\in\mathcal S}
[
\mathbf v_s^{(l)},
\operatorname{mean}(\mathbf v_s^{(l)})
].
\]

For angular calibration, the entropy statistic is

\[
\mathcal H_\theta^{(l)}
=
-
\frac{
\sum_n p_\theta^{(l)}(n)\log p_\theta^{(l)}(n)}
{\log M}.
\]

The angular descriptor concatenates layer-wise entropy and final-box
anisotropy:

\[
\mathbf u_A=
[
\mathcal H_\theta^{(1)},\ldots,
\mathcal H_\theta^{(L)},
a^{(L)}
].
\]

A lightweight MLP is then used to calibrate the classification logit.

\section{Implementation and Training Details}
\label{app:implementation}

\begin{table*}[!t]
\caption{Hyperparameter configurations for RiO-DETRv2 models.}
\vspace{-0.4cm}
\begin{center}
\resizebox{\textwidth}{!}{
\begin{tabular}{l|ccccc}
\toprule
\textbf{Setting}
& \textbf{RiO-DETRv2-x}
& \textbf{RiO-DETRv2-l}
& \textbf{RiO-DETRv2-m}
& \textbf{RiO-DETRv2-s}
& \textbf{RiO-DETRv2-n} \\
\midrule

Backbone Name
& HGNetv2-B5 & HGNetv2-B4 & HGNetv2-B2
& HGNetv2-B0 & HGNetv2-B0 \\
Input Resolution
& $1024^2$ & $1024^2$ & $1024^2$ & $1024^2$ & $1024^2$ \\
Embedding Dimension
& 384 & 256 & 256 & 224 & 128 \\
Feedforward Dimension
& 2048 & 1024 & 1024 & 1024 & 512 \\
Decoder Layers
& 4 & 4 & 3 & 3 & 3 \\
Queries
& 300 & 300 & 300 & 300 & 300 \\
Denoising Queries
& 100 & 100 & 100 & 100 & 100 \\
Number of Levels
& 3 & 3 & 3 & 3 & 2 \\
Boundary Bins
& 33 & 33 & 33 & 33 & 33 \\
Boundary Regression Scale
& 8.0 & 4.0 & 4.0 & 4.0 & 4.0 \\
Angle Bins
& 36 & 36 & 36 & 36 & 36 \\
Angular Refinement Factor
& 1.5 & 1.5 & 1.5 & 1.5 & 1.5 \\
Angle-Aware LQE
& Enabled & Enabled & Enabled & Enabled & Enabled \\
\midrule

Optimizer
& AdamW & AdamW & AdamW & AdamW & AdamW \\
Base LR
& $8{\times}10^{-5}$ & $8{\times}10^{-5}$
& $1{\times}10^{-4}$ & $1{\times}10^{-4}$
& $2{\times}10^{-4}$ \\
Backbone LR
& $8{\times}10^{-6}$ & $8{\times}10^{-6}$
& $1{\times}10^{-5}$ & $1{\times}10^{-5}$
& $2{\times}10^{-5}$ \\
Weight Decay
& $1.25{\times}10^{-4}$ & $1.25{\times}10^{-4}$
& $1{\times}10^{-4}$ & $1{\times}10^{-4}$
& $1{\times}10^{-4}$ \\
\midrule

Weight of $\mathcal{L}_{\mathrm{VFL}}$
& 1 & 1 & 1 & 1 & 1 \\
Weight of $\mathcal{L}_{\mathrm{L1}}$
& 5 & 5 & 5 & 5 & 5 \\
Weight of $\mathcal{L}_{\mathrm{KLD}}$
& 5 & 5 & 5 & 5 & 5 \\
Weight of $\mathcal{L}_{\mathrm{FGL}}$
& 0.15 & 0.15 & 0.15 & 0.15 & 0.15 \\
Weight of $\mathcal{L}_{\mathrm{A\text{-}FGL}}$
& 0.15 & 0.15 & 0.15 & 0.15 & 0.15 \\
Weight of $\mathcal{L}_{\mathrm{DDF}}$
& 1.5 & 1.5 & 1.5 & 1.5 & 1.5 \\
Weight of $\mathcal{L}_{\mathrm{Distill}}$
& 20 & 20 & 20 & 20 & 20 \\
\midrule

Total Batch Size
& 16 & 16 & 16 & 16 & 16 \\
Total Epochs
& 72 & 102 & 132 & 142 & 160 \\
Flat Epochs
& 20 & 51 & 41 & 70 & 80 \\
No Aug Epochs
& 12 & 12 & 12 & 12 & 12 \\
\bottomrule
\end{tabular}
}
\vspace{-0.5cm}
\end{center}
\label{tab:riov2_hyperparameters}
\end{table*}

\subsection{Method-specific Hyperparameters}

Table~\ref{tab:riov2_hyperparameters} summarizes the hyperparameters
introduced by RiO-DETRv2. In particular, for AI-TOD-R, we use 900 queries for all models instead of the default 300.

\subsection{Latency Measurement}

For fairness, all models are evaluated for single-image end-to-end model inference latency on a single NVIDIA T4 GPU with TensorRT. Here, end-to-end model inference refers to the complete GPU inference pipeline from the network input to the final detection outputs. All measurements use a batch size of one.

\textbf{Engine execution.} Each model is serialized into a TensorRT engine and executed via \texttt{IExecutionContext}. For TensorRT 10 engines, inference is launched with \texttt{execute\_async\_v3}; otherwise, we fall back to \texttt{execute\_async\_v2}.

\textbf{I/O shapes and memory binding.} For dynamic-shape engines, we select optimization profile 0 and benchmark at its OPT shape. GPU buffers for all inputs and outputs are allocated once as CUDA tensors, and their device pointers are bound to TensorRT using tensor addresses for TensorRT 10 or bindings for legacy versions. This avoids repeated memory-allocation overhead and ensures that the benchmark measures the GPU inference pipeline itself.

\textbf{Post-processing.} For detectors that require post-processing, such as non-end-to-end models, rotated NMS is implemented using a TensorRT NMS plugin and is included in the reported latency. End-to-end detectors do not use NMS or other external detection post-processing. This protocol measures each model according to its actual inference pipeline and therefore provides a more faithful comparison of practical inference speed.

\textbf{Evaluation scope.} The reported latency is the inference time for a single input image or image patch. Dataset-level preprocessing and post-processing operations, such as cropping a large DOTA image into multiple patches and merging the patch-level predictions, are not included. Image loading, host-to-device data transfer, and other CPU-side preprocessing operations are also excluded.

\textbf{Warm-up.} We run 100 warm-up iterations before timing to stabilize the GPU state and eliminate one-time overhead, such as CUDA context initialization. All warm-up iterations are executed asynchronously on a dedicated CUDA stream, followed by synchronization.

\textbf{Timing protocol.} Latency is measured using CUDA events recorded on the same CUDA stream. We record a start event, execute inference for 300 iterations back-to-back, record an end event, and synchronize on the end event. The average per-image latency in milliseconds is computed as
\begin{equation}
\mathrm{Latency} = \frac{t_{\mathrm{elapsed}}}{N},
\end{equation}
where $t_{\mathrm{elapsed}}$ is the elapsed time measured by CUDA events in milliseconds and $N$ is the number of inference runs. Throughput is reported as
\begin{equation}
\mathrm{FPS} = \frac{1000}{\mathrm{Latency}}.
\end{equation}

\begin{figure*}[t]
\centering
\includegraphics[width=0.93\textwidth]{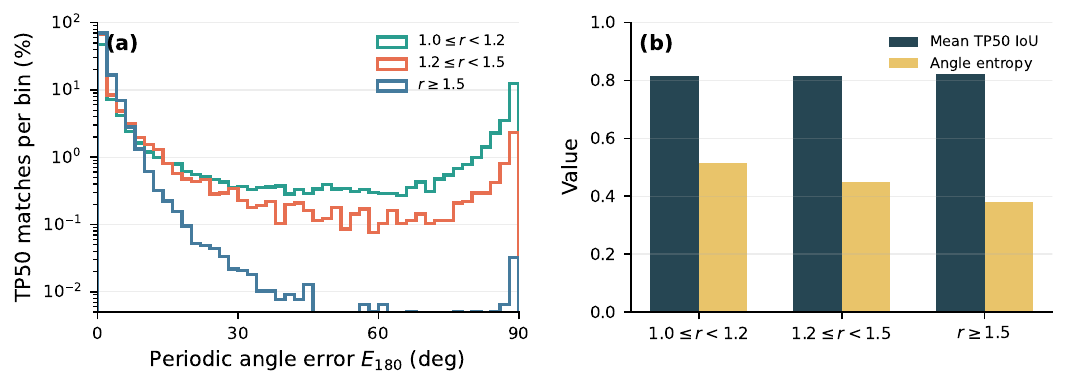}
\caption{Behavior across GT aspect-ratio groups on DIOR-R. (a) The standard
angular error develops a pronounced secondary mode near $90^\circ$ for
near-square objects, directly exposing occasional canonical-axis exchange.
(b) Despite this parameterization ambiguity, matched-box IoU remains nearly
unchanged, while angular distribution entropy increases toward the square
limit.}
\label{fig:near_square_analysis}
\end{figure*}

\begin{figure*}[t]
\centering
\includegraphics[width=\linewidth]
{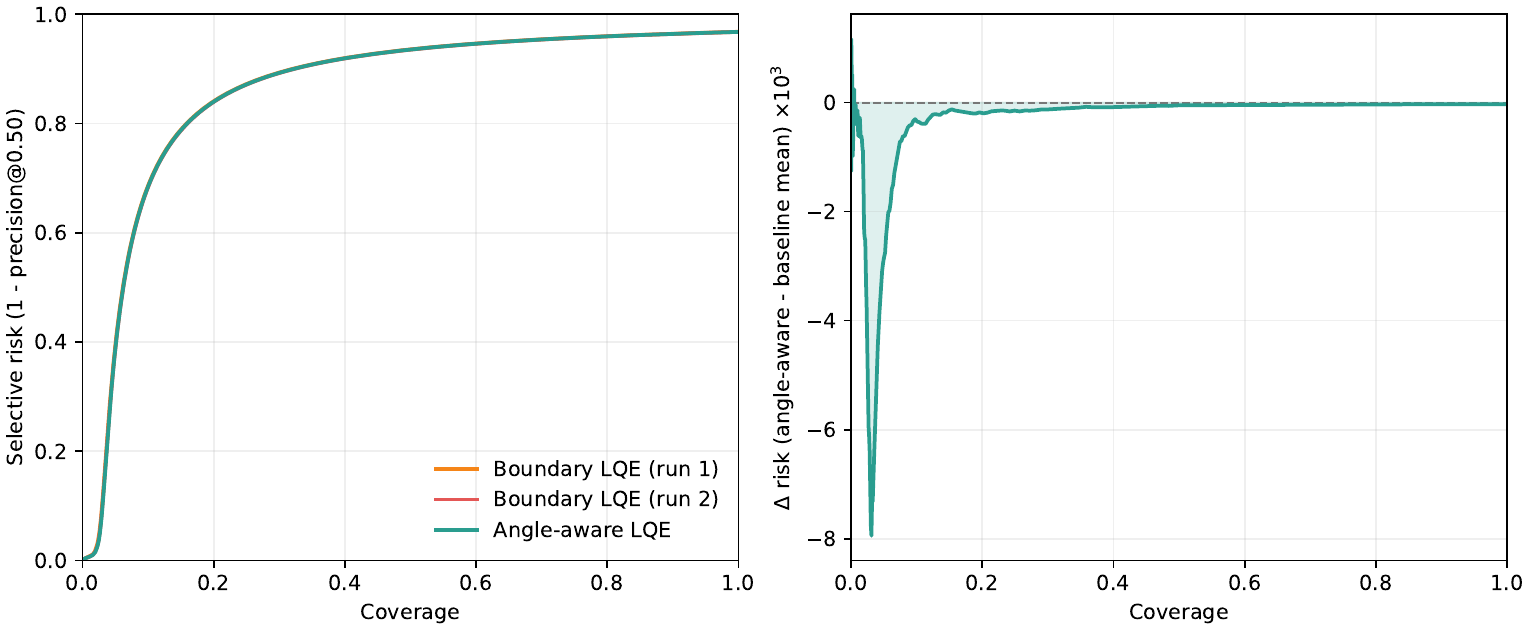}
\caption{Risk--coverage analysis over all top-300 predictions on DIOR-R. The
left panel presents the complete risk--coverage curves, and the right panel
shows the risk difference between the angle-aware estimator and the mean of
the two boundary-only runs.}
\label{fig:all_prediction_risk_coverage}
\end{figure*}

\section{Empirical Analysis of Distributional Uncertainty}
\label{sec:appendix_uncertainty}

We provide a comprehensive empirical analysis of the uncertainty information
encoded by the predicted boundary and angular distributions. A meaningful
distributional uncertainty signal should satisfy three properties: it should
correlate with the realized localization error, support effective ranking of
predictions by reliability, and improve the alignment between detection
confidence and complete OBB quality. We evaluate these properties using
pretrained RiO-DETRv2-$x$ on the complete DIOR-R test set.

Our analysis consists of two complementary settings. We first study the
correspondence between distribution concentration and localization error on
matched detections, which directly evaluates the geometric information encoded
by the distributions. We then evaluate confidence calibration over all
predictions retained by the standard top-300 postprocessor, including unmatched
predictions, duplicate detections, classification errors, and inaccurately
localized predictions.

\paragraph{Distribution-error correspondence.}
For each image, the top 300 predictions are matched to ground-truth instances
using class-aware one-to-one matching at an OBB IoU threshold of 0.5, resulting
in 114,229 matched detections. This setting enables a direct instance-wise
comparison between the predicted distribution shape and the realized
localization error.

For boundary localization, we average the normalized entropies of the four
final-layer object-aligned boundary distributions and denote the result by
$H_{\mathrm{bnd}}$. We also examine their peak probabilities, which directly
characterize distribution concentration and constitute the principal
statistics used by the boundary quality estimator. For orientation, we use
both the final-layer entropy $H_{\theta}^{L}$ and the mean entropy over all
decoder layers $\bar{H}_{\theta}$. Because the orientation of a nearly square
OBB is intrinsically ambiguous, angular statistics are evaluated on objects
with aspect ratio at least 1.5, leaving 78,448 matched detections. Localization
risk is measured by $1-\mathrm{IoU}_{\mathrm{OBB}}$, while angular risk is
measured by the shortest periodic angular error.

\begin{table*}[t]
\centering
\caption{Relationship between distributional uncertainty and localization
error on matched DIOR-R detections. Spearman correlation is reported with a
95\% image-level bootstrap confidence interval. ``10\% retained'' denotes the
10\% predictions with the lowest estimated uncertainty.}
\label{tab:uncertainty_correlation}
\setlength{\tabcolsep}{4.5pt}
\resizebox{0.6\linewidth}{!}{%
\begin{tabular}{lccc}
\hline
\textbf{Uncertainty signal} &
\textbf{Spearman $\rho$} &
\textbf{10\% retained} &
\textbf{All matched} \\
\hline
$H_{\mathrm{bnd}}$ vs.\ $1-\mathrm{IoU}_{\mathrm{OBB}}$ &
$0.782\,[0.774,0.788]$ & $0.049$ & $0.180$ \\
$\bar{H}_{\theta}$ vs.\ angle error &
$0.672\,[0.662,0.682]$ & $0.240^\circ$ & $1.877^\circ$ \\
$H_{\theta}^{L}$ vs.\ angle error &
$0.685\,[0.674,0.696]$ & $0.237^\circ$ & $1.877^\circ$ \\
\hline
\end{tabular}
}
\end{table*}

\begin{table*}[t]
\caption{Detection and orientation statistics on DIOR-R grouped by GT aspect
ratio. R@50 and R@75 use independent class-aware one-to-one matching.
$E_{\mathrm{axis}}$ is a diagnostic measure for near-$90^\circ$ axis exchange,
not a replacement for the standard angular error.}
\label{tab:near_square_analysis}
\centering
\resizebox{\textwidth}{!}{
\begin{tabular}{lrrrrrrrr}
\toprule
\textbf{GT aspect ratio} & \textbf{\#GT} & \textbf{R@50 (\%)} &
\textbf{R@75 (\%)} & \textbf{Mean IoU} & \textbf{$E_{180}$ MAE} &
\textbf{$E_{\mathrm{axis}}$ MAE} & \textbf{Axis switch (\%)} &
\textbf{Angle entropy} \\
\midrule
$1.0\leq r<1.2$ & 27,035 & 92.64 & 72.26 & 0.817 & $25.71^\circ$ &
$5.29^\circ$ & 20.69\% & 0.515 \\
$1.2\leq r<1.5$ & 11,521 & 91.53 & 71.30 & 0.814 & $7.83^\circ$ &
$3.46^\circ$ & 4.14\% & 0.451 \\
$r\geq1.5$ & 85,882 & 90.62 & 72.68 & 0.822 & $1.87^\circ$ &
$1.81^\circ$ & 0.05\% & 0.380 \\
\bottomrule
\end{tabular}
}
\end{table*}

\paragraph{Correlation with localization error.}
As shown in Table~\ref{tab:uncertainty_correlation}, boundary entropy exhibits
a strong positive rank correlation of $0.782$ with complete OBB localization
error. Conversely, the mean boundary peak probability has a correlation of
$-0.763$ with $1-\mathrm{IoU}_{\mathrm{OBB}}$, indicating that concentrated
boundary distributions are associated with more accurate oriented boxes.

The angular distributions exhibit similar behavior. The mean and final-layer
angular entropies achieve correlations of $0.672$ and $0.685$, respectively,
with the realized angular error. Likewise, one minus the final angular peak
probability has a correlation of $0.684$ with angular error. The agreement
between entropy- and peak-based measures indicates that distribution
concentration provides a stable instance-wise indicator of localization
difficulty.

\begin{figure*}[t]
\centering
\includegraphics[width=0.49\linewidth]
{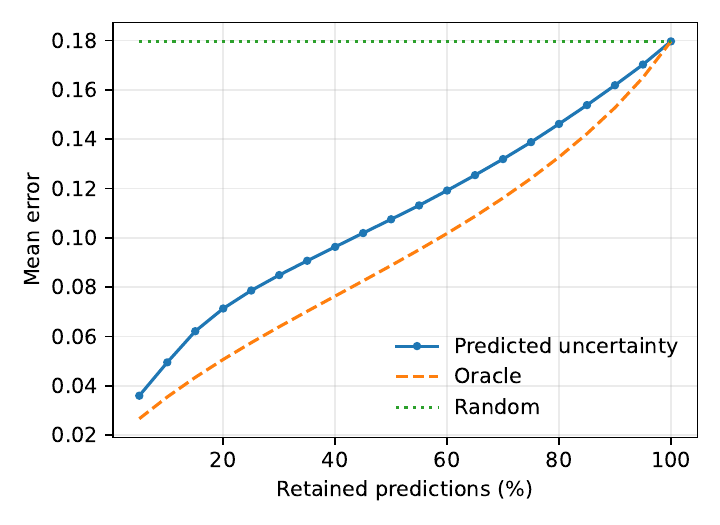}
\hfill
\includegraphics[width=0.49\linewidth]
{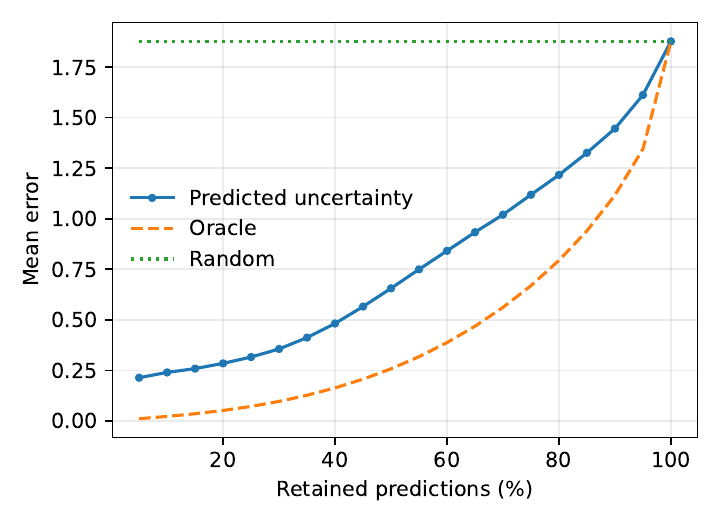}
\caption{Error-retention curves on DIOR-R. Predictions are retained in
ascending order of estimated uncertainty. Left: complete OBB localization risk
ranked by boundary entropy. Right: angular error ranked by mean angular
entropy. Lower curves indicate more effective uncertainty ranking.}
\label{fig:uncertainty_retention}
\end{figure*}

\paragraph{Error-retention analysis.}
Figure~\ref{fig:uncertainty_retention} further evaluates whether the
distributional statistics can identify reliable predictions. Retaining the
10\% matched detections with the lowest boundary entropy reduces mean
$1-\mathrm{IoU}_{\mathrm{OBB}}$ from $0.180$ to $0.049$, corresponding to a
$72.5\%$ reduction. Similarly, retaining the 10\% lowest-angular-entropy
detections reduces mean angular error from $1.877^\circ$ to $0.240^\circ$, a
reduction of $87.2\%$. Both retention curves remain substantially below random
retention, demonstrating that the predicted distribution shapes provide
effective rankings of localization reliability.

\paragraph{Probability assigned to the regression target.}
We additionally measure the soft-target negative log-likelihood (NLL), where
the continuous regression target is interpolated between its two neighboring
bins using the same target construction as in training. The boundary
distributions obtain an NLL of $2.434$, substantially below the
uniform-distribution reference $\log(33)=3.497$. The final angular distribution
obtains an NLL of $1.677$, compared with the uniform reference
$\log(36)=3.584$. Their corresponding top-1-bin calibration errors are
$0.0057$ and $0.0208$, respectively. These results demonstrate that the
predicted distributions assign meaningful probability mass around the
realized regression targets, rather than merely serving as an alternative
parameterization for coordinate regression.

\paragraph{Calibration over all retained predictions.}
We further evaluate whether distribution-derived reliability improves the
final detection confidence over the complete prediction set. For every image,
we retain all 300 outputs without applying any score, class-correctness, or IoU
filter, resulting in 3,521,400 predictions per model. Predictions are processed
in descending confidence order and matched to ground truth using class-aware
one-to-one OBB matching. At threshold $\tau$, the correctness target is defined
as
\begin{equation}
y_{\tau} =
\mathbb{I}\!\left[
c=\hat{c},\,
\mathrm{IoU}_{\mathrm{OBB}}\geq\tau,\,
\text{GT unmatched}
\right].
\end{equation}
This evaluation therefore jointly accounts for classification correctness,
localization accuracy, duplicate detections, and false positives.

We report expected calibration error (ECE) using 15 equal-width bins, binary
NLL, Brier score, and the area under the risk--coverage curve (AURC). To
directly evaluate confidence alignment with complete OBB quality, we
additionally define
\begin{equation}
q_{50} =
\begin{cases}
\mathrm{IoU}_{\mathrm{OBB}}, & y_{0.5}=1,\\
0, & \text{otherwise},
\end{cases}
\end{equation}
and report quality-aware ECE (Q-ECE) and the Spearman correlation between
confidence and $q_{50}$.

\begin{table*}[t]
\centering
\caption{Calibration over all top-300 predictions on DIOR-R. The two
boundary-only rows correspond to independent training runs. The angle-aware
estimator introduces only 113 additional parameters. Lower is better for ECE,
NLL, Brier, AURC, and Q-ECE.}
\label{tab:all_prediction_calibration}
\setlength{\tabcolsep}{3.5pt}
\resizebox{0.7\linewidth}{!}{%
\begin{tabular}{lcccccc}
\hline
\textbf{Score estimator} &
\textbf{ECE$_{50}$} &
\textbf{NLL$_{50}$} &
\textbf{Brier$_{50}$} &
\textbf{AURC$_{50}$} &
\textbf{Q-ECE} &
\textbf{$\rho(\mathrm{conf.},q_{50})$} \\
\hline
Boundary LQE (run 1) &
$0.0494$ & $0.0698$ & $0.0113$ & $0.8607$ & $0.0552$ & $0.3011$ \\
Boundary LQE (run 2) &
$0.0490$ & $0.0691$ & $0.0114$ & $0.8596$ & $0.0548$ & $0.3025$ \\
Angle-aware LQE &
$\mathbf{0.0455}$ & $\mathbf{0.0649}$ & $\mathbf{0.0106}$ &
$0.8599$ & $\mathbf{0.0514}$ & $0.3021$ \\
\hline
\end{tabular}
}
\end{table*}

Relative to the mean of the two independently trained boundary-only models,
the angle-aware estimator reduces ECE$_{50}$ by $7.4\%$, NLL$_{50}$ by
$6.5\%$, Brier score by $6.4\%$, and Q-ECE by $6.7\%$. Paired image-level
bootstrap gives 95\% confidence intervals of $[-0.0041,-0.0036]$ and
$[-0.0037,-0.0032]$ for the ECE$_{50}$ difference against the two
boundary-only runs, respectively, confirming that the improvement is
consistent across test images.

The improvement is also preserved under a stricter localization criterion. At
$\tau=0.75$, the angle-aware model obtains an ECE of $0.0521$, compared with
$0.0556$--$0.0561$ for the boundary-only models, while NLL decreases from
$0.0812$--$0.0816$ to $0.0777$. Meanwhile, AURC and confidence--quality rank
correlation remain competitive with the boundary-only models. These results
show that angle-aware LQE improves absolute confidence calibration while
preserving the underlying prediction ranking.

Figure~\ref{fig:all_prediction_risk_coverage} presents the corresponding
risk--coverage behavior. The angle-aware estimator maintains consistently
competitive selective risk across the complete coverage range, demonstrating
that the calibration improvement is obtained without compromising confidence
ordering.

\paragraph{Discussion.}
The results provide consistent evidence that the predicted distributions encode
meaningful localization uncertainty. First, boundary and angular entropy show
strong instance-wise correlations with their corresponding geometric errors,
and peak-probability statistics exhibit consistent complementary behavior.
Second, uncertainty-based retention substantially reduces both complete OBB
localization risk and angular error. Third, the distributions assign
significantly higher likelihood to the realized regression targets than a
uniform reference. Finally, incorporating angular distribution statistics into
LQE consistently reduces ECE, NLL, Brier score, and quality-aware calibration
error over the complete prediction set.

Taken together, these results demonstrate that the discrete distributions are
not merely a regression representation. Their concentration provides an
effective estimate of localization reliability, and the proposed quality
estimators successfully transfer this information to the final classification
confidence. This establishes a direct empirical connection between
distributional uncertainty, complete OBB quality, and confidence calibration.

\section{Analysis of Near-Square Objects}
\label{app:near_square}

\paragraph{Motivation and protocol.}
Long-edge canonicalization assigns the local horizontal axis to the major axis
of an OBB, providing consistent side semantics for anisotropic objects.
However, as the aspect ratio approaches one, the major axis becomes weakly
identifiable and a small geometric perturbation can exchange the two axes,
causing an approximately $90^\circ$ parameterization switch. We study this
degenerate regime using the predictions of RiO-DETRv2-x on the complete
DIOR-R test set, which contains 11,738 images and 124,438 annotated objects.
For each GT box, we define $r=\max(w,h)/\min(w,h)$ and divide the objects into
$[1.0,1.2)$, $[1.2,1.5)$, and $[1.5,\infty)$. Predictions are matched to GTs
class-wise and in descending confidence order, independently at IoU
thresholds 0.5 and 0.75.

For every TP at IoU 0.5, we report the standard $\pi$-periodic angular error
$E_{180}$. To diagnose coordinate-axis exchange without replacing the standard
angular metric, we additionally define
\begin{equation}
E_{\mathrm{axis}}
=
\min\left(E_{180},\left|90^\circ-E_{180}\right|\right).
\label{eq:axis_equiv_error}
\end{equation}
An axis switch is counted when $E_{180}>45^\circ$ and
$E_{\mathrm{axis}}<10^\circ$. The angular entropy is the normalized
distribution entropy averaged over all decoder layers.

\paragraph{Results and discussion.}
Table~\ref{tab:near_square_analysis} shows that long-edge canonicalization is
indeed discontinuous near the square limit. In the $1.0\leq r<1.2$ group, the
standard angular MAE rises to $25.71^\circ$ and the detected axis-switch rate
reaches 20.69\%. Nevertheless, the same group has an axis-equivalent angular
MAE of only $5.29^\circ$. The secondary mode close to $90^\circ$ in
Fig.~\ref{fig:near_square_analysis}(a) further confirms that much of the large
standard angular error is caused by canonical-axis exchange rather than an
arbitrary orientation failure.

Importantly, this representation ambiguity has limited impact on geometric
detection quality. Across all three groups, mean matched IoU varies by only
0.008 and R@75 remains between 71.30 and 72.68. Meanwhile, angular entropy
increases monotonically from 0.380 for elongated objects to 0.515 for
near-square objects, indicating that the predicted distributions expose the
weak orientation identifiability of this regime. These observations refine
the scope of the long-edge representation: its object-aligned side semantics
are stable away from the square degeneracy; near that limit, the canonical
frame can switch, but the resulting ambiguity is largely parameterization
level, preserves OBB geometry, and is reflected by higher angular uncertainty.

\end{document}